\documentclass{article}
\usepackage{times}
\usepackage{soul}
\usepackage{url}
\usepackage[hidelinks]{hyperref}
\usepackage[utf8]{inputenc}
\usepackage[small]{caption}
\usepackage{graphicx}
\usepackage{amsmath}
\usepackage{amsthm}
\usepackage{booktabs}
\usepackage{algorithm}
\usepackage{algorithmic}
\usepackage{subcaption}

\usepackage{amssymb}

\title{Eliminating Backdoor Triggers for Deep Neural Networks Using Attention Relation Graph Distillation}

 \author{
 Jun Xia,
 Ting Wang,
 Jieping Ding,
 Xian Wei,
 Mingsong Chen$^*$\\
 $^1$Shanghai Key Lab of Trustworthy Computing, East China Normal University\\
$^*$Corresponding Author, \{mschen\}@sei.ecnu.edu.cn
}

\begin{document}

\maketitle

\begin{abstract}
Due to the prosperity of Artificial Intelligence (AI) techniques, more and more
backdoors are designed by adversaries
to attack Deep Neural Networks (DNNs).
Although the state-of-the-art method Neural 
Attention Distillation (NAD)
 can effectively erase backdoor 
triggers from DNNs, 
it still suffers from non-negligible Attack Success Rate (ASR) together with 
lowered classification ACCuracy (ACC), since
NAD focuses on backdoor defense using 
attention features (i.e., attention maps) 
of the same order.
In this paper, we   
 introduce  a novel backdoor defense framework named  Attention Relation Graph Distillation (ARGD), which fully  explores the correlation among attention features  with different orders using our proposed 
 Attention Relation  Graphs (ARGs).
Based on the 
alignment of 
ARGs between both teacher and student models during knowledge 
 distillation,  
 ARGD can  eradicate more backdoor triggers than NAD. 
Comprehensive experimental results show 
 that, against six latest backdoor attacks, ARGD outperforms 
NAD by up to 94.85\% reduction in ASR, while  ACC can be improved by up to 3.23\%. 
\end{abstract}

\section{Introduction}

\label{sec:intro}
Along with the 
proliferation of Artificial Intelligence (AI) techniques, 
 Deep Neural Networks (DNNs)  are increasingly
 deployed in various safety-critical domains, e.g., autonomous driving, commercial surveillance, and medical monitoring. 
Although DNNs enable both intelligent sensing and control, 
more and more of them are becoming the main target of adversaries.
It is reported that DNNs are prone to be attacked 
by potential threats 
in different phases of their 
life cycles \cite{Song2021FDA3FD}. For example, 
due to biased training data or overfitting/underfitting models, 
at test time a tiny
input perturbation made by some adversarial 
attack can fool a given
DNN and  result in 
incorrect or unexpected
behaviors \cite{evaluation}, which may cause disastrous consequences.
As another type of  notoriously perilous adversaries,  backdoor attacks can inject   
triggers in DNNs  on numerous occasions, e.g., 
collecting training data from unreliable sources, and 
downloading pre-trained DNNs from untrusted parties. 
Typically, by poisoning a small portion of  training data,  backdoor attacks aim to trick DNNs 
into learning the correlation between 
trigger patterns and  target labels. 
Rather than affecting
the  performance of models on clean data, 
backdoor attacks may
cause incorrect
prediction at test time when some trigger
pattern appears \cite{backdoor-federated,Backdoor_phy}.

Compared with traditional adversarial attacks, 
backdoor attacks have gained more attentions, since they can be easily implemented in 
real scenarios \cite{blend,badnets}.
Currently, there are two major kinds of mainstream backdoor defense methods. The first one is the 
detection-based  methods that can 
identify whether there exists a backdoor attack during
the training process. Although these approaches are promising in 
preventing  DNNs from  backdoor attacks, they cannot 
fix models implanted 
with backdoor triggers. 
The second one is  
 the erasing-based methods, which  aims to eliminate backdoor triggers by purifying the 
malicious impacts of backdoored models.
In this paper, we focus on the latter case.
Note that, due to the concealment and imperceptibility of backdoors, it is hard to fully purify backdoored DNNs. Therefore, our goal is to further lower  
Attack Success Ratio (ASR) on backdoored data without 
sacrificing the classification 
ACCuracy (ACC) on clean data.

Neural Attention Distillation (NAD)
\cite{NAD} has been recognized as
the most effective backdoor erasing method so far, which is implemented
based on  finetuning and distillation
operations. 
Inspired by the concept of 
attention transfer \cite{komodakis2017paying}, NAD utilizes 
a teacher model to guide the finetuning of  a backdoored student model using a small set of clean data. Note that  the teacher model is obtained by finetuning the student model using the same set of clean data. By aligning intermediate-layer attention features of  the student model
with their counterparts in the teacher model,  backdoor triggers can be effectively erased from DNNs. 
In NAD, 
an attention  feature represents
the activation information 
of all neurons in one layer. Therefore, the conjunction of 
all the feature  attentions  within a DNN
can reflect the most discriminative regions in the model's topology \cite{lopez}.

Although the attention mechanism can be used as an indicator to evaluate
the performance of backdoor erasing methods, 
the  implementation of NAD strongly limits the  expressive power of attention features, since it only compares the feature 
attentions of the same order during the finetuning. Unfortunately,  
the correlation among  attention features of different orders \cite{liu2019knowledge,interpreting}
is totally ignored. 
The omission of such salient  
features in finetuning may
result in  a ``cliff-like'' decline in   defending backdoor attacks \cite{komodakis2017paying}. 
 In this paper, we  propose a novel backdoor erasing framework named  Attention Relation Graph Distillation (ARGD), which fully  considers
 the correlation of attention features of different  orders. 
 This paper  makes the following three major contributions:
\begin{itemize}
\item
We propose  Attention Relation Graphs (ARGs) to fully reflect  the correlations 
among attention features of different orders, which can be 
combined with  distillation to erase more  backdoor
triggers from DNNs. 
\item
We define three loss functions for ARGD, which enable effective 
alignment of the intermediate-layer ARG of a student model with that of its teacher model. 
\item
We conduct comprehensive experiments on various well-known backdoor attacks to show the
effectiveness and  efficiency  of our proposed defense method. 
\end{itemize}

The rest of this paper is organized as follows. After the 
introduction to related work on backdoor attack and defence methods 
in Section~\ref{sec:related_work}, Section~\ref{sec:approach} details our ARGD approach. Section \ref{sec:exp} presents the experimental results on well-known benchmarks under six state-of-the-art backdoor attacks. Finally, Section \ref{sec:con} concludes the paper.


\section{Related Work}
\label{sec:related_work}
{\bf Backdoor Attacks:}
We are witnessing more and more DNN-based backdoor attacks in real environment \cite{backdoor-federated,Adi2018TurningYW}. Typically, 
a backdoor attack refers to 
designing a trigger pattern injected into partial 
training data with (poisoned-label attack \cite{badnets}) or without (clean-label attack \cite{liu2020reflection}) a target label.
At  test time,  
such backdoor patterns can be triggered 
to control 
the prediction results, which may result in incorrect or unexpected behaviors. 
Aiming at increasing ASR without affecting ACC, 
extensive studies \cite{Li2020RethinkingTT} have been investigated 
to design specific backdoor triggers.
Existing backdoor attacks can be 
classified into two categories, i.e.,   observable backdoor attacks, and  imperceptible backdoor attacks  \cite{cleanlabel}.
Although the observable backdoor attacks have a profound impact on DNNs,  the training data with changes by such attacks can be 
easily  identified. As an alternative, the imperceptible backdoor attacks (e.g., natural reflection\cite{liu2020reflection} and human imperceptible noises \cite{haoti}) are  more commonly used in practice. 


{\bf Backdoor Defense:}
The mainstream backdoor defense approaches
can be classified into two major types. 
The first one is the detection-based 
methods, which can identify  
backdoor triggers from DNNs
 during the training  \cite{backdoor-detect} or   filtering  backdoored training data to eliminate the influence of backdoor attacks \cite{Chou2020SentiNetDL}.
Note that
few of existing detection-based methods
can be used to  purify  backdoored DNNs.
The second one is 
the  elimination-based approaches \cite{neuralclean,goldblum2020adversarially,peiself2021}.
Based on a limited number of clean data, 
such methods can erase backdoor triggers by  
finetuning the
 backdoored DNNs. 
 Although  various  
 elimination-based  approaches \cite{NAD,MCR}
 have bee extensively 
 investigated, 
 so far there is no method that can fully purify the backdoored DNNs.
Most  of them 
are still striving to improve ASR and 
ACC from different perspectives. 
For example, the Neural Attention Distillation (NAD) method adopts attention features
of the same order
to improve   
backdoor elimination  performance based on finetuning and  distillation operations.  
However, NAD  suffers from non-negligible ASR. This is because NAD 
focuses on the alignment of feature 
attentions of the same order, thus 
the expressive power of attention features is inevitably limited.

To the best of our knowledge, ARGD is the first attempt that takes the correlation of 
attention features into account for the purpose of eliminating backdoor triggers from DNNs. Based on our proposed ARGs and corresponding loss functions, ARGD 
can not only reduce the ASR significantly, but also improve the ACC on clean data.

\section{Our ARGD Approach}
\label{sec:approach}

As the state-of-the-art elimination-based backdoor 
defense method, NAD tries to  suppress
the impacts  of backdoor attacks based on model  retraining (finetuning) and 
knowledge distillation 
of  backdoored models. Based on clean retraining data, NAD
 can effectively erase backdoor triggers by aligning the 
 intermediate-layer attention features
 between teacher and student models. 
However,  due to the privacy issues or various access restrictions, in practice
such clean data  for finetuning only accounts for  a very small proportion of the data required for model training. This  strongly limits the 
defense performance of NAD, since NAD focuses on
the alignment of attention features of the same orders, while the  relation of transforms 
between attention features is totally ignored. As a result of limited retraining data, 
it is hard to guarantee the ASR and ACC performance for NAD.

\begin{figure}[h]
	\vspace{-0.1in}
 	\begin{center}
 		\includegraphics[width=\linewidth]{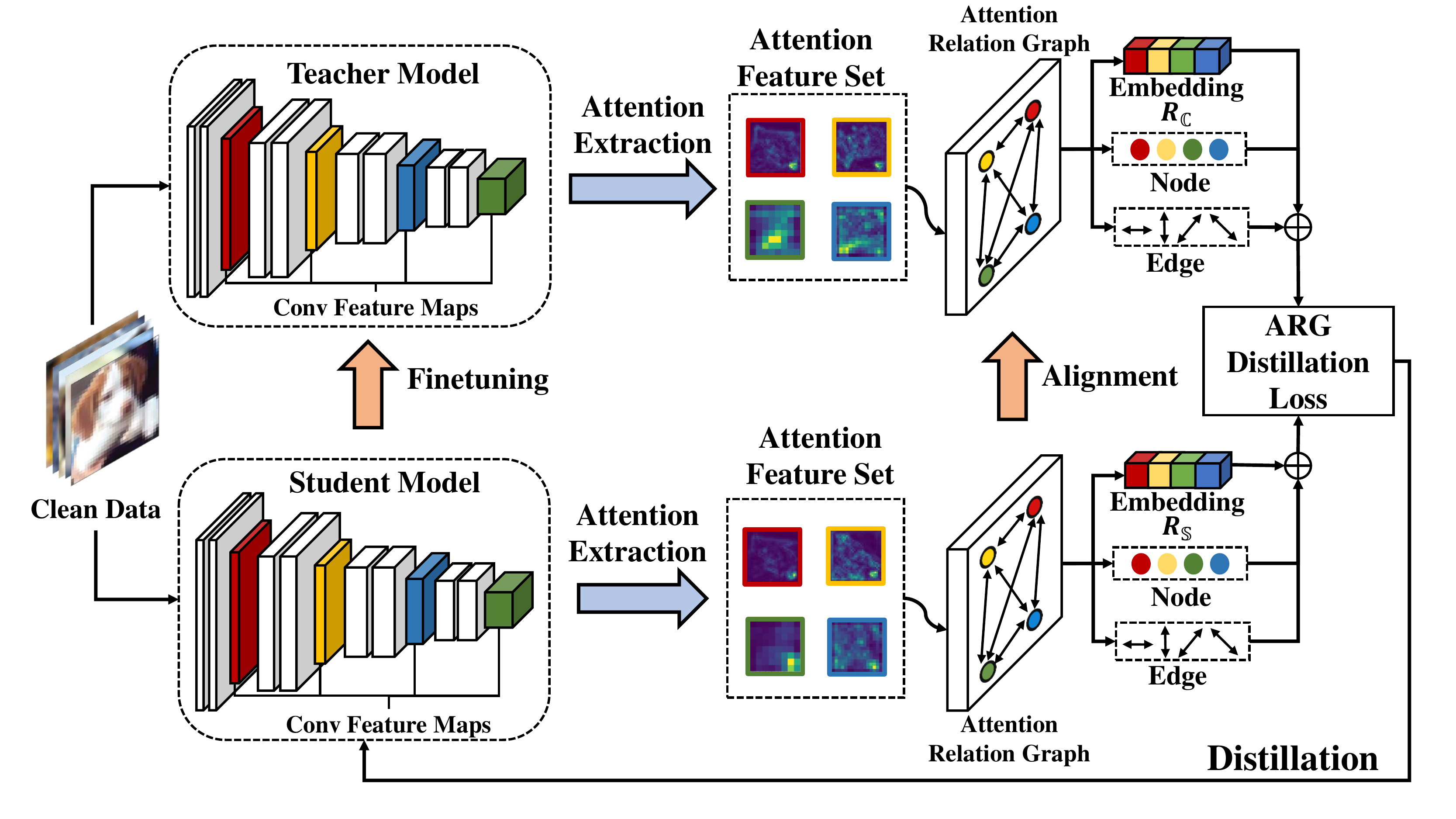}
 		\vspace{-0.15in}
 		\caption{Overview of attention relation  graph distillation} 
 		\label{fig:ARGD}
 		\vspace{-0.1in}
 	\end{center}
\end{figure}

To address the ASR and ACC issues posed by NAD,  
we introduce a novel knowledge distillation method named ARGD
  as shown in Figure~\ref{fig:ARGD}, which 
  fully considers the correlations between attention features
using our proposed ARGs for backdoor defense. 
This figure has two parts, where the upper part denotes both the teacher model 
and its extracted ARG information. The teacher model is
trained by the finetuning of the backdoored student model using the 
provided clean data. 
The lower part of the figure presents the student model, which needs to be finetuned 
by aligning its ARG to  the one of the teacher model. We use the ARG distillation loss 
for knowledge distillation, which takes the combination of  node, edge and
embedding  correlations  into account. 
The following subsections will introduce the key components of our approach in detail.

\subsection{Attention Relation Graph}

Inspired by the  
instance relation graph introduced  in 
\cite{liu2019knowledge}, we propose ARGs to enable  the modeling of 
knowledge transformation relation between attention features and facilitate
the alignment of defense structures against backdoor triggers from 
 student models to  teacher models. Unlike instance relation graphs
 that are established based on the regression accuracy of  image instances, for 
 a given input data, 
 an ARG  of  is built on top of the model's
 attention features within different orders. 
In our approach, we assume that the finetuned teacher model by clean data has a benign knowledge structure represented by its ARGs, which fully reflects the correlations between its attention features of different orders. Therefore, we use  ARGs to guide the finetuning of backdoored student model during the 
knowledge distillation by aligning the ARGs of the 
backdoored student model to its counterparts of the teacher model. Given an
input data, the ARG of a model  can be modeled as a complete graph formalized by a 2-tuple $G=(\mathbb{N}, \varepsilon)$, where 
 $\mathbb{N}$ represents the node set and $\varepsilon$ denotes the edge set. Here, each node in $\mathbb{N}$
 represents an attention feature with a specific order, and 
 each edge in $\varepsilon$ indicates the 
 similarity between two nodes.

\subsubsection{ARG Nodes}

Given a DNN model $M$ and an input data $X$, we define the $p^{th}$ convolutional feature map of $M$  as $F^{p} = M^{p}(X)$, which is an activation map having the
three dimensions of
 channel index, width  and height.
  By taking the
 3-dimensional
 $F^{p}$ as an input, the attention extraction operation
 $\mathcal{A}$ outputs a flattened 2-dimensional
 tensor $T^{p}_M$
 representing the extracted attention feature.
 Let  $C$, $H$, $W$ denote
 the  number of channels,  height, and  width of  input tensors, 
 respectively. Essentially, the attention extraction operation
 can be formulated as a function 
 $\mathcal{A}_M: \mathbb{R}^{C \times H \times W} \rightarrow \mathbb{R}^{ H \times W}$
 defined as follows:
 \begin{footnotesize} 
\begin{equation}
\footnotesize
\mathcal{A}_M (F^{p}) = \frac{1}{C}\sum_{i=1}^{C} \left | F^{p}_{i}(X) \right |^{2},
\nonumber
\label{node_distance}
\end{equation}
\end{footnotesize} 
where $C$ is the number of channels of  $F^{p}$, and  $F^{p}_i$ indicates the $i^{th}$ channel of $F^{p}$.
By applying $\mathcal{A}_M$ on $F^{p}$, we can obtain the attention 
feature of $F^{p}$, which is denoted as an ARG node with an order of 
$p$. Assuming that the model $M$ has 
$k$ convolutional feature maps, based on $\mathcal{A}_M$ we can construct 
a node set $\mathbb{N} = \left \{ T^{1}_{M}, T^{2}_{M}, ..., T^{p}_{M},...,T^{k}_{M}\right \}$. Note that  in practice we only use a subset of $\mathbb{N}$
to construct ARGs.


\subsubsection{ARG Edges}

After figuring out the node set to construct an ARG, we need to 
construct a complete graph, where the edge set (i.e., $\varepsilon = \bigcup_{i=1}^{k}\bigcup_{j=1}^{k}\left \{e^{i j}\right \}$) indicates the correlations
between attention features of different orders in $M$, where 
$e^{i j}$ indicates the edge between $T_M^i$ and $T_M^j$.
Let $E_M$ be an weight 
function of edges in the form
of $E_M:\varepsilon \rightarrow \mathbb{R}$, where 
$E_M^{i j}=E_M(e^{i j})$ denotes the Euclidean distance between  two  attention features $T_M^i$ and $T_M^j$. Assume that the maximum 
size of  $T_M^i$ and $T_M^j$ is $h\times w$.
Let $\Gamma_{ij}(Y)$ be a function
that converts
the attention feature $Y$ into a 2-dimensional 
feature $Y^\prime$ with a size of $h\times w$.
 $E_M$  indicates
the correlations  between  attention features, where the edge weight $E^{i j}$ can be  calculated as
\begin{footnotesize} 
\begin{equation}
E_M^{i j} = \| \Gamma_{ij}({T_M^{i}}) - \Gamma_{ij}({T_M^{j}})  \|_2.
\nonumber
\label{edge_distance}
\end{equation}
\end{footnotesize} 

\subsection{ARG Embedding}

To facilitate the alignment from a student  ARG to 
its teacher counterpart,  we consider the graph embedding for ARGs, 
where an ARG embedding can be constructed by all the involved attention features within a model. 
Since the embedding
  reflects high-dimensional semantic features of all the  nodes in an ARG, they can be used to figure out the 
  knowledge dependencies between ARGs of both the 
  teacher and student models. 
  Let $\mathbb{C}$ and $\mathbb{S}$
be the teacher model and student model, respectively. We construct  ARG embedding vectors (i.e.,  $R^{p}_{\mathbb{C}}$ and $R^{p}_{\mathbb{S}}$) from  the $p^{th}$ attention features of 
 $\mathbb{C}$ and $\mathbb{S}$, respectively,
 based on the following two formulas:
\begin{equation}
 \begin{matrix}
R^{p}_{\mathbb{C}}  = \sigma (W^{p}_{\mathbb{C}}\cdot\psi(T^{p}_{\mathbb{C}})), \
R^{p}_{\mathbb{S}}  = \sigma (W^{p}_{\mathbb{S} }\cdot\psi(T^{p}_{\mathbb{S} })), 
\end{matrix}
\nonumber
\label{global_pool}
\end{equation}
where $\psi(\cdot)$ is the adaptive average pooling function, and $\sigma(\cdot)$ is the activation function to generate the embedding vectors. 
Here, $W^{p}_{\mathbb{C}}$ and $W^{p}_{S}$ are two  linear transformation parameters constructed 
in the distillation process for  
the $p^{th}$ attentions feature of  the teacher and student models.


 By comparing the embedding vectors between the teacher model and the student model, 
 we can figure out the correlation between a student  node and all the 
teacher  nodes. In our approach, we use 
the relation vector $\beta^{p}_{\mathbb{\mathbb{S}}}$ to denote the 
correlations between the $p^{th}$ student node and 
all the teacher nodes, which is defined as
\begin{footnotesize} 
\begin{equation}
\beta^{p}_{\mathbb{\mathbb{S}}} = \textit{Softmax}(R^{p}_{\mathbb{S}} {^{\mathsf{T}}} \cdot w^{b}_{1}\cdot R^{1}_{\mathbb{C} }, \ldots, R^{p}_{\mathbb{S}} {^{\mathsf{T}}}
\cdot w^{b}_{p}\cdot R^{p}_{\mathbb{C} }, \ldots, R^{p}_{\mathbb{S}}  
{^{\mathsf{T}}}
\cdot w^{b}_{k}\cdot R^{k}_{\mathbb{C} } ),
\nonumber
\label{embedding_distance}
\end{equation}
\end{footnotesize} 
where  $w^{b}$ is the bilinear weight used to convert the underlying relation between different order attention features in distillation \cite{pirsiavash2009bilinear}.

\subsection{ARG Distillation Loss}

The ARG distillation loss $\mathfrak{L}_{G}$ is defined as the difference between ARGs. It involves three kinds of differences from different perspectives 
between the teacher ARG $G_{\mathbb{C}}$ and  student ARG $G_{\mathbb{S}}$: 
 i)  node difference that indicates the sum of distances between  node pairs in terms of attention features; ii) edge difference that 
specifies the sum of distances between  edge pairs; and iii) 
embedding difference that denotes the 
weighted sum of distances between  student-teacher node pairs in terms  of embedding vectors. To reflect such differences from different  structural perspectives, we define
 three kinds of losses, i.e., 
 ARG node loss $\mathfrak{L}_{\mathbb{N}}$,  ARG edge loss $\mathfrak{L}_{\varepsilon}$ and ARG embedding loss $\mathfrak{L}_{Em}$.
 Since the weight of an ARG edge indicates the similarity between two nodes with different orders, the ARG edge loss can further enhance the alignment of ARGs between the teacher model and student model. 
 The ARG node loss function is defined as 
 \begin{footnotesize} 
\begin{equation}
    \mathfrak{L}_{\mathbb{N}}\left ( \mathbb{N}_{\mathbb{S}}, \mathbb{N}_{\mathbb{C}} \right ) = \frac{1}{k} \sum_{i=0}^{k} {\left \| \frac{T_{\mathbb{C}}^{i}}{\left \|T_{\mathbb{C}}^{i} \right \|_{2}} - \frac{T_{\mathbb{S}}^{i}}{\left \|T_{\mathbb{S}}^{i} \right \|_{2}} \right \|_{2}}.
    \nonumber
    \label{node_loss}
\end{equation}
\end{footnotesize} 
The ARG node loss $\mathfrak{L}_{N}$ is essentially a kind of imitation loss, which enables the  
pixel-level alignment of attention features at same layers
from a backdoored student model  to its teacher counterpart.
The ARG edge loss denotes the difference between two edge sets, 
which is calculated using
\begin{footnotesize} 
\begin{equation}
    \mathfrak{L}_{\varepsilon}\left ( E_{\mathbb{S}}, E_{\mathbb{C}} \right )= \frac{1}{\mathcal{C}_{k}^{2}}\sum_{i=1}^{k-1}\sum_{j=i+1}^{k}\left \|E_{\mathbb{C}}^{ij} - E_{\mathbb{S}}^{ij}  \right \|_{2}^{2},
    \nonumber
\end{equation}
\end{footnotesize} 
where $\mathcal{C}_{k}^{2}$ is the combination formula. During the alignment of ARGs, 
an attention feature of the student model needs to learn
 knowledge from different attention features  of the  teacher model.  However,  the combination 
 of ARG node loss and edge loss cannot fully explore the 
 knowledge structure 
 dependence among   attention features
 between the 
 teacher model and 
 student model. To 
  enable such kind of learning, we propose the 
ARG embedding loss based on the relation vector,  
which is defined as 
\begin{equation}
     \mathfrak{L}_{Em}\left ( T_{\mathbb{C}}, T_{\mathbb{S}} \right ) = \sum_{i=1}^{k} \sum_{j=1}^{k} \beta^{i, j}_{\mathbb{S}}\left \| \Gamma_{ij} (T^{i}_{\mathbb{C}})  -\Gamma_{ij}( T^{j}_{\mathbb{S}}) \right \|_{2}.
     \nonumber
\end{equation}
 Based on the above three losses, 
 we define the ARG distillation loss $\mathfrak{L}_{G}$  
 to support accurate ARG alignment during the knowledge distillation, which is defined as
 \begin{footnotesize} 
\begin{equation}
    \mathfrak{L}_{G}\left ( G_{\mathbb{S}}, G_{\mathbb{C}} \right )= \mathfrak{L}_{\mathbb{N}}+ \mathfrak{L}_{\varepsilon} + \mathfrak{L}_{Em}.
    \nonumber
    \label{dis_loss}
\end{equation}
\end{footnotesize}



\subsection{Overall Loss for Distillation}
Our ARGD method is based on knowledge distillation. To enable the alignment of ARGs 
during the distillation process, we define the overall loss function of the backdoored DNN as
\begin{equation}
    \mathfrak{L}_{overall}=\mathfrak{L}_{CE}+ \mathfrak{L}_{G},
    \nonumber
\end{equation}
where $\mathfrak{L}_{CE}$ is the cross entropy loss between  predictions of the
backdoored DNN and corresponding
target values.

\section{Experimental Results}
\label{sec:exp}

To evaluate the effectiveness of  our approach,  we implemented our ARGD framework on top of Pytorch (version 1.4.0). All the experiments were conducted on a workstation with Ubuntu operating system, Intel i9-9700K CPU, 16GB memory, and NVIDIA GeForce GTX2080Ti GPU.
In this section, we designed comprehensive
experiments to answer the following three research questions.

\textbf{Q1 (Superiority of ARGD)}: 
What are the advantages of ARGD compared with state-of-the-art methods?

\textbf{Q2 (Applicability of ARGD)}: 
What are the impacts of 
 different settings (e.g.,  clean data rates,  teacher model architectures) on the performance of ARGD?
 
\textbf{Q3 (Benefits of ARGs)}: 
Why our proposed 
ARGs can substantially improve purifying 
backdoored DNNs?


\subsection{Experimental Settings}

{\bf Backdoor Attacks and Configurations:}
We conducted  experiments using the following six latest
backdoor attacks: i) BadNets \cite{badnets}, ii) Trojan attack \cite{trojan}, iii) Blend attack \cite{blend}, iv)
Sinusoidal signal attack (SIG) \cite{sig},  v)  Clean Label \cite{cleanlabel}, and vi) Reflection attack (Refool) \cite{liu2020reflection}.
To make a fair comparison against these methods, we adopted the same  configurations (e.g., backdoor trigger patterns, backdoor trigger sizes, and target labels for 
restoring) as presented in their original papers.
Based on WideResNet (WRN-16-1) \cite{resnet}
and its variants, we  trained  DNN models based on 
the CIFAR-10 dataset using  
our approach and its six opponents, respectively. Note  that here 
each DNN training for backdoor attacks involves 100 epochs. 


\begin{table*}[h]
\footnotesize
\centering 
\resizebox{\linewidth}{!}{
\begin{tabular}{c||cc|cc|cc|cc|cc||cc}

\hline
Backdoor          & \multicolumn{2}{c|}{\begin{tabular}[c]{@{}c@{}}Backdoored\end{tabular}} & \multicolumn{2}{c|}{Finetuning}                    & \multicolumn{2}{c|}{MCR (t=0.3)}                    & \multicolumn{2}{c|}{NAD}                           & \multicolumn{2}{c||}{ARGD (Ours) }    &   \multicolumn{2}{c}{Improvement}                \\
Attack            & \multicolumn{1}{c}{ASR(\%)}                    & \multicolumn{1}{c|}{ACC(\%)}            & \multicolumn{1}{c}{ASR(\%)} & \multicolumn{1}{c|}{ACC(\%)} & \multicolumn{1}{c}{ASR(\%)} & \multicolumn{1}{c|}{ACC(\%)} & \multicolumn{1}{c}{ASR(\%)} & \multicolumn{1}{c|}{ACC(\%)} & \multicolumn{1}{c}{ASR(\%)} & \multicolumn{1}{c||}{ACC(\%)} & \multicolumn{1}{c}{ASR(\%)} & \multicolumn{1}{c}{ACC(\%)}\\ \hline
BadNets           & 100.00                                     & 80.08                               & 4.56                    & 77.16                    & 3.12                    & 78.99                    & 3.29                    & 77.98                    & \textbf{2.10}           & \textbf{79.81}     & 41.99 &  2.35   \\
Trojan            & 99.81                                      & 80.04                               & 3.57                    & 78.06                    & 2.56                    & 77.76                    & 2.91                    & 77.03                    & \textbf{1.97}           & \textbf{79.60}  &32.30 &  \textbf{3.23}       \\
Blend             & 79.42                                      & 82.76                               & 3.08                    & 80.08                    & 70.06                   & 77.10                    & 2.33                    & 79.09                    & \textbf{0.12}           & \textbf{80.47}   & \textbf{94.85} &  1.74        \\
SIG & 99.98                                      & 82.43                               & 9.12                    & 79.08                    & 3.69                    & 81.52                    & 11.78                    & 79.63                  & \textbf{1.83}           & \textbf{80.56}       & 84.47 &  1.17     \\
Clean Label       & 45.94                                      & 82.43                               & 11.42                   & 81.24                    & 16.56                   & 79.25                    & 9.56                    & 79.66                    & \textbf{5.32}           & \textbf{80.18}   & 45.14 &  2.98     \\
Refool            & 100.00                                     & 82.22                               & 5.96                    & 80.23                    & 8.94                    & 79.99                    & 4.02                    & 80.87                    & \textbf{3.12}           & \textbf{81.67}    & 22.39 &  0.99       \\ \hline
Average           & 87.53                                      & 81.66                              & 6.29                    & 79.31                    & 17.49                   & 79.10                    & 5.70                    & 79.04                   & \textbf{2.41}           & \textbf{80.38} & +53.52 &   +2.08        \\ 
Deviation          & \multicolumn{1}{c}{-}             & \multicolumn{1}{c|}{-}              & -81.24                  & -2.35                    & -70.04                   & -2.56                    & -81.82                    & -2.29                      & \textbf{-85.12}          & \textbf{-1.38} & \multicolumn{1}{c}{-} & \multicolumn{1}{c}{-}           \\\hline
\end{tabular}
  }
  \caption{Performance 
  of 4 backdoor defense methods against 6 backdoor attacks. The deviations indicate  the percentage changes in average
  ASR/ACC compared to the baseline {\it Backdoored}. The best experimental results in ASR and ACC are marked in bold.}    \label{tab_1}
  \vspace{-0.15in}
\end{table*}


{\bf Defense Method Settings and Evaluation:}
We compared our ARGD with three state-of-the-art backdoor defense methods, i.e.,  traditional finetuning \cite{papernot2016distillation}, Mode Connectivity Repair (MCR) \cite{MCR}, and  NAD \cite{NAD}. 
Since it is difficult to achieve clean data for the purpose of finetuning in practice, similar to the work presented in \cite{NAD}, in our experiments we assumed that all the 
defense methods can  access only 5\% of training dataset as the clean dataset by default.
We conducted the image preprocessing using the same training configuration 
of NAD adopted  in \cite{NAD}.
We set the mini-batch size of all the defense
methods to 64, and the  
initial learning rate 
to 0.1. 
For each backdoor defense method, we trained 
each DNN for 10 epochs for the purpose of erasing 
backdoor triggers. We adopted the  Stochastic Gradient Descent (SGD) optimizer  with a momentum of 0.9.
Similar to the setting of attack  model training, by default we use WideResNet (WRN-16-1) as the
teacher model of ARGD for finetuning.  
However, it does not mean that the structures of 
both student and teacher models should be the same. 
In fact, teacher models with different structures can
also be applied on ARGD (see Table~\ref{tab_3} for more details). During the finetuning, based on the attention extraction operation,
 our approach can extract
attention features 
of each group of the WideResNet model and form an 
ARG for the given DNN. 
We use two indicators to evaluate the 
performance of  backdoor defense methods: i) 
Attack Success Rate (ASR) denoting the 
 ratio of succeeded attacks over all the attacks
 on backdoored data; and ii)
 the  classification ACCuracy  (ACC) 
 indicating the ratio of correctly predicted 
 data over all the clean data. 
 Generally, lower  ASRs mean    better  
defense capabilities.

\subsection{Comparison with State-of-the-Arts}

To show the superiority of  ARGD, we compared 
our approach with the three   backdoor 
defense methods  against six latest  backdoor attacks.
Table~\ref{tab_1} presents the comparison results. 
Column 1 presents the name of six backdoor attack methods. 
Column 2 shows the  results for 
 backdoored student models without any defense. Column 3 gives the results for the finetuning methods. Note that here the finetuning method was conducted based on the counterpart
teacher model 
with extra 10 epoch training on the same collected clean data.  
 Columns 4-6 denote the experimental results for 
 MCR, NAD and ARGD, respectively. Column  7 shows the improvements of ARGD 
 over NAD for the six backdoor attacks.

From this table, we can find that ARGD can not 
only purify the backdoored DNNs effectively, 
but also have the minimum side effect on clean data. 
We can  observe that, among all the four defense methods, 
ARGD outperforms the other 
 three defense methods significantly. 
Especially, ARGD greatly outperforms the state-of-the-art approach NAD from the perspectives  of both ASR and ACC.
As shown in the last column, compared with NAD, ARGD can reduce the ASR by up to 94.85\%   and increase the ACC by up to 3.23\%.
The reason of such improvement is mainly because ARGD takes the alignment of 
ARGs into account during the finetuning between teacher and student models, 
while NAD only considers
the attention features of the same order during the finetuning.
Without considering the structural information of ARGs, the 
finetuning using attention features can be easily biased, which  
 limits the backdoor erasing 
capacities of attention features as well as degrades the ACC on clean data.

\subsection{Impact of Clean Data Sizes}

Since the finetuning is mainly based on the learning on clean data, the clean data sizes play an important role in determining the 
quality of backdoor defense. Intuitively,
the more clean data we can access for finetuning, the better ASR and ACC we can  achieve. 
Table~\ref{tab_2} presents the performance of the four  defense methods against the six backdoor attack approaches under different clean data sizes. Due to  space limitation, 
this table only shows the averaged ASR and ACC values of
the six backdoor attack methods. 
In this table, column 1 presents the clean data size information in terms of clean data ratio. 
Here, we investigated different ratios from 1\% to 20\% of the total training data. For example, 
5\% means that we use 5\% of the original clean training data for the finetuning between teacher and student 
models. Column 2 presents the averaged ASR and ACC values for all the backdoored DNNs using
the testing data, and columns 3-6 show the ASR and ACC for the four defense methods, respectively. The last column denotes
the improvement of ARGD over NAD.

From this table, we can find that 
  ARGD has the best performance
  in eliminating backdoor triggers. 
  Compared with {\it Backdoored},  
  ARGD can reduce ASR by up to 2.41\% from 
  87.53\%, while  the finetuning method  and NAD 
   reduce ASR by up to 4.38\% and 3.91\%, respectively.
 Among all the four cases, our approach can  achieve the highest ACC in three out of four cases. 
 Especially,  ARGD outperforms both the finetuning method and NAD
 in all the cases from
   the perspectives of both ASR and ACC. For example,
   when the ratio of clean data is 1\%, 
   ARGD outperforms NAD by 43.89\% and 19.53\% 
   for ASR and ACC, respectively. 
   Note that, when the
   clean data ratio is  1\%, ARGD can achieve an ASR of  3.58\%, which is much 
  smaller than   all the cases of the other three defense methods with different clean data ratios. 
  It means
  that the backdoor erasing effect of ARGD with  only 1\% clean data   can achieve much better   ASR than the other three methods 
  with 20\% clean data each.  
 For the case with 1\% clean data ratio, although MCR 
 can have a slightly higher ACC than ARGD, 
 its ASR is much higher than the other three defense methods. 
 This implies that MCR has a higher dependence on clean data and is more prone to attacks when there are little  clean data for finetuning.


\begin{table*}[h]
\centering 
\footnotesize
\resizebox{\linewidth}{!}{
\begin{tabular}{c||cc|cc|cc|cc|cc||cc}
\hline
\multicolumn{1}{c||}{Clean Data} & \multicolumn{2}{c|}{\begin{tabular}[c]{@{}c@{}}Backdoored \end{tabular}} & \multicolumn{2}{c|}{Finetuning}               & \multicolumn{2}{c|}{MCR (t=0.3)}               & \multicolumn{2}{c|}{NAD}               & \multicolumn{2}{c||}{ARGD (Ours)}  & \multicolumn{2}{c}{Improvement}\\
\multicolumn{1}{c||}{Ratio(\%)} & ASR(\%)                     & \multicolumn{1}{c|}{ACC(\%)}                      & ASR(\%) & \multicolumn{1}{c|}{ACC(\%)}        & ASR(\%) & \multicolumn{1}{c|}{ACC(\%) }       & ASR(\%) & \multicolumn{1}{c|}{ACC(\%)}  & ASR(\%) & \multicolumn{1}{c||}{ACC(\%)} & ASR(\%)        & ACC(\%)      \\ \hline
\multicolumn{1}{c||}{1}         & 87.53                        & \multicolumn{1}{c|}{81.66 }                        & 7.78    & \multicolumn{1}{c|}{76.04}          & 41.34   & \multicolumn{1}{c|}{\textbf{79.88}} & 6.38   & \multicolumn{1}{c|}{64.06}   & \textbf{3.58}  & \multicolumn{1}{c||}{76.57}   &  43.89 &      \textbf{19.53}  \\
\multicolumn{1}{c||}{5}         & 87.53                        & \multicolumn{1}{c|}{81.66 }                        & 6.29    & \multicolumn{1}{c|}{79.31}          & 17.49   & \multicolumn{1}{c|}{79.10}          & 5.70    & \multicolumn{1}{c|}{79.04}   & \textbf{2.41}  & \multicolumn{1}{c||}{\textbf{80.38}}  & \textbf{57.72} &      1.70\\
\multicolumn{1}{c||}{10}        & 87.53                        & \multicolumn{1}{c|}{81.66 }                        & 6.66    & \multicolumn{1}{c|}{80.75}          & 14.21     & \multicolumn{1}{c|}{80.29}          & 5.18    & \multicolumn{1}{c|}{80.69}   & \textbf{3.01}  & \multicolumn{1}{c||}{\textbf{81.21}}  &41.89  & 0.64 \\
\multicolumn{1}{c||}{20}        & 87.53                        & \multicolumn{1}{c|}{81.66 }                        & 4.38    & \multicolumn{1}{c|}{82.17} & 7.01    & \multicolumn{1}{c|}{82.06}          & 3.91   & \multicolumn{1}{c|}{82.31}   & \textbf{2.64}  & \multicolumn{1}{c||}{\textbf{82.52}}    &32.48  & 0.26      \\ \hline
\end{tabular}
}
\vspace{-0.1in}
  \caption{Performance of 4 backdoor defense methods against 6  backdoor attacks under different clean data ratios.}
    \label{tab_2}
    \vspace{-0.1in}
\end{table*}

\subsection{Impact of Teacher Model Architectures}

\begin{table*}[]
\centering 
\footnotesize
\resizebox{\linewidth}{!}{
\begin{tabular}{c||cc|c|cc|cc|cc||cc}
\hline
Model          & Teacher   & Student   & Teacher & \multicolumn{2}{c|}{\begin{tabular}[c]{@{}c@{}}Backdoored \end{tabular}} & \multicolumn{2}{c|}{NAD} & \multicolumn{2}{c||}{ARGD (Ours)} & \multicolumn{2}{c}{Improvement} \\
Difference     & Structure & Structure & ACC(\%) & ASR(\%)                                 & ACC(\%)                                & ASR(\%)     & ACC(\%)    & ASR(\%)        & ACC(\%)   & ASR(\%) & ACC(\%)      \\ \hline
Same Model          & WRN-16-1  & WRN-16-1  & 67.51   & 45.94                                   & 82.43                                    & 6.16        & 64.76     & \textbf{4.84}  & \textbf{74.02} & 21.43 & 14.30 \\
Depth          & WRN-10-1  & WRN-16-1  & 62.31   & 45.94                                   & 82.43                                  & 5.96        & 60.46      & \textbf{4.55}  & \textbf{70.78}  & 23.66 & \textbf{17.07}\\

Channel        & WRN-16-2  & WRN-16-1  & 68.93   & 45.94                                   & 82.43                                   & 7.98        & 66.63      & \textbf{5.46}  & \textbf{76.11} & 31.58 & 14.28\\
Depth \& Channel & WRN-40-2  & WRN-16-1  &69.01   & 45.94                                   & 82.43                                    & 8.08        & 67.15      & \textbf{4.92}  & \textbf{76.45} &\textbf{39.11} & 13.85\\ \hline
\end{tabular}
  }
 \vspace{-0.1in}
  \caption{Performance  of 2 distillation-based backdoor defense methods against  Clean Label  attacks  with different teacher models.}
  \label{tab_3}
 \vspace{-0.15in}
\end{table*}


In knowledge distillation, the performance of  student models is mainly
determined by the knowledge level of  teacher models.
However,  due to the uncertainty and unpredictability of training processes, 
it is hard to figure out an ideal 
teacher model for specific student models for the purpose of
backdoor defense.
Rather than exploring
optimal teacher models, in this experiment we 
investigated the impact of teacher model architectures on the backdoor
defense performance. Due to space limitation, here we only consider the 
case of Clean Label backdoor attacks. 


Table~\ref{tab_3} presents the results of defense performance comparison  between 
NAD and  ARGD. For both methods, we considered four 
different teacher model architectures
denoted by ``{\it WRN-x-y}'', where $x$ and $y$ indicate the depth
of convolutional layers and the model channel width of a WideResNet, respectively. 
The first column presents the differences between pairs of teacher and 
student models. Column 2 shows the architecture settings for both teacher and student models. Based on the teacher models trained using the 
5\% clean training data, column 3 gives the prediction results on all the provided
testing data in CIFAR-10. Column 4 presents the ASR and ACC information for the 
backdoored student models, which are the same as the ones shown in 
Table~\ref{tab_1}. Columns 5-6 denote the 
defense performance of both NAD and ARGD methods. The last column 
indicates the improvements of ARGD over NAD.

From this table, we can find  that  model architectures with 
larger depths or channel widths can
lead to better accuracy as shown in column 3.  
This is also true for the ACC results of  both NAD and ARGD methods.  
Since ASR and ACC are two conflicting targets for backdoor defense,
we can observe that larger teacher models will result in the 
reverse trends for  ASR. Note that, no matter what the teacher model architecture is,   ARGD always outperforms NAD for both ASR and ACC. 
For example, when we adopt a teacher model with architecture
WRN-10-1, ARGD can improve the ASR and ACC  of NAD by 23.66\% and 17.07\%,
respectively.

\subsection{Understanding  Attention Relation Graphs}

\begin{figure}[h]
 \vspace{-0.1in}
\centering
\footnotesize
\includegraphics[width=.8\linewidth]{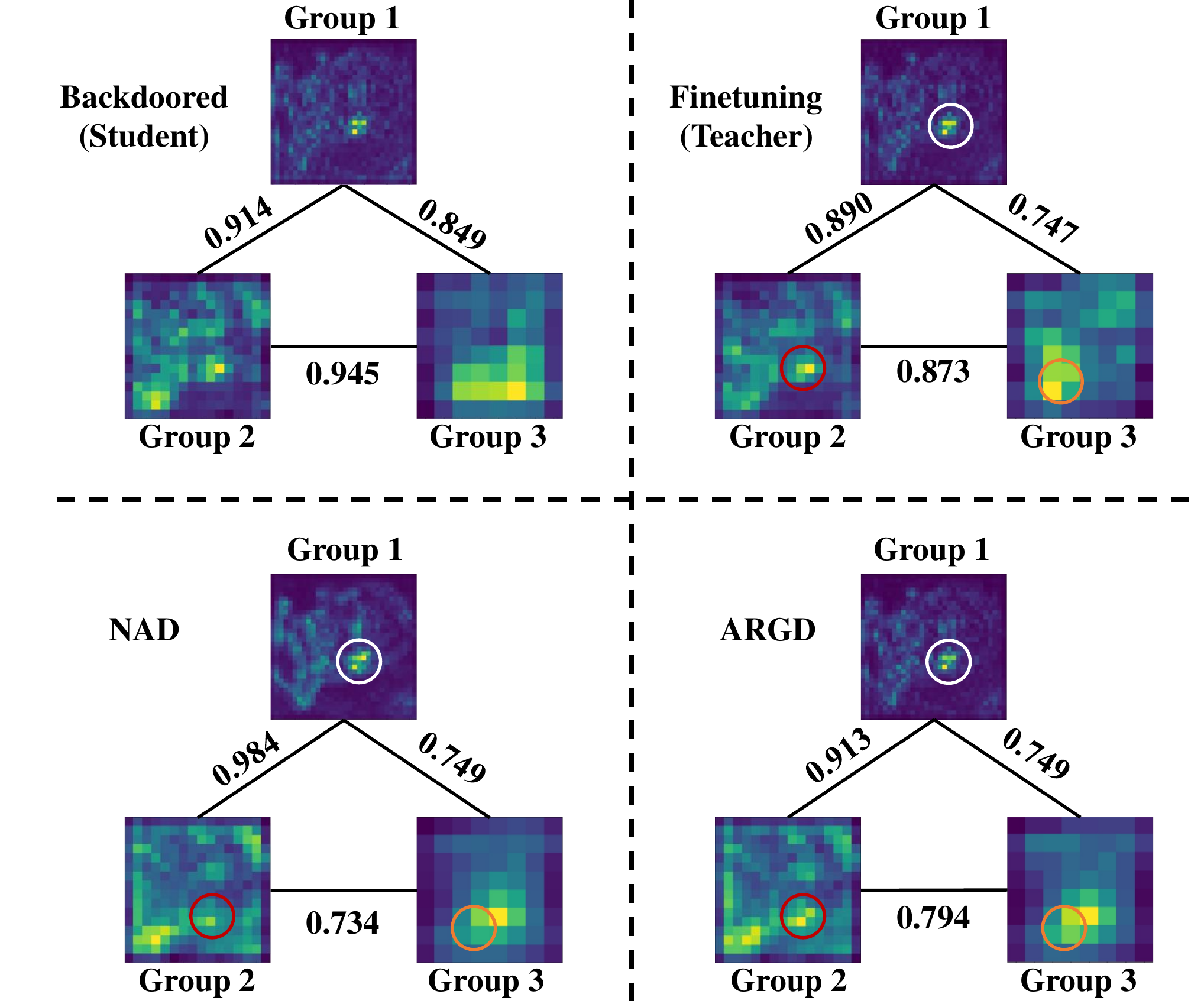}
 \vspace{-0.1in}
\caption{Visualization of ARGs generated by
different defense methods for a BadNets backdoored image. The two ARGs at bottom are
generated by the student models of 
 NAD and ARGD.} 
\vspace{-0.15in}
\label{vis for ARGD}
\end{figure}

To  understand how ARGs help  eliminating backdoor triggers, Figure~\ref{vis for ARGD} presents
a comparison  of ARGs generated  by different defense 
methods for a BadNets backdoored image. 
Since both teacher and student models used by the involved  defense methods are based on model WRN-16-1
that has three residual groups, each ARG here 
has three nodes representing attention features, where  
the  lighter color indicates higher attention values. 
In this figure, the student models of NAD and ARGD are learnt
based on the knowledge distillation using the backdoored student model and finetuning teacher model with the 5\% clean training data. In the finetuning teacher model, we used 
circles with specific  colors to  highlight the most noticeable areas in different
ARG nodes, respectively. Similarly, 
to enable similarity analysis of student models, we also labeled the circles with the same sizes, colors and locations on the ARG nodes of NAD and ARGD.

From this figure, we can observe
that,
benefiting from the imitative learning of
ARGs, our proposed ARGD method can achieve better ARG 
alignment between the teacher model and 
student model than the one of NAD. Compared with NAD,
ARGD  can not only  generate closer attention 
features with different orders  (especially the part inside the circle  of group 2) 
for its student model, but also have closer correlation between 
attention features. 
For example, the correlations between 
the attention feature pairs of (group1, group2)
and (group2, group3) are 0.913 and 0.794, 
while the corresponding  correlations for the ARG 
generated by NAD are  0.984 and 0.734, respectively. 
Since the edge weights of the finetuning teacher model are 0.890 and 0.873, respectively,  ARDG has better alignment than 
NAD for these two ARG edges. 
In other words, by using ARG-based 
 knowledge transfer, 
  the effects of backdoor triggers
  can be effectively 
suppressed, while the  benign
knowledge structure is minimally affected.





\begin{table}[h]
\vspace{-0.1in}
\scriptsize
\centering
\begin{tabular}{cccc|cc}
\hline
Finetuning & Node & Edge & Embedding & ACC (\%)   & ASR (\%)   \\ \hline
$\checkmark$                 &      &      &           & 79.31 & 6.29  \\
$\checkmark$                    & $\checkmark$      &      &           & 79.04 & 5.70  \\
$\checkmark$                    & $\checkmark$      & $\checkmark$      &           & 79.88 & 3.03  \\
$\checkmark$                    & $\checkmark$      & $\checkmark$      & $\checkmark$           & \textbf{80.38} & \textbf{2.41} \\ \hline
\multicolumn{4}{c|}{Backdoored DNN}     & 81.66 & 87.53 \\ \hline
\end{tabular}
  \vspace{-0.1in}
  \caption{Ablation  results considering  impacts of 
  ARG components.}
  \label{Ablation}
  \vspace{-0.15in}
\end{table}

To evaluate the contributions of key ARG components in ARGD, we conducted a  series of ablation studies, whose results 
are shown in 
Table~\ref{Ablation}. Column 1 denotes the
case without adopting knowledge distillation or 
incorporating any of our proposed loss functions.  Based on our 
ARGD method, columns 2-4 present the three cases indicating
whether the node, edge and embedding losses
are included, respectively. Columns 5-6
indicate the average ACC and ASR  of   the
  six backdoor attacks under  5\% clean training data, respectively.
The last row specifies the average ACC and ASR results for
the backdoored DNNs without any defense. 
Note that NAD can be considered as ARGD with 
only the node loss. 
Compared with the finetuning method, 
the ASR of NAD can be improved from 6.29\% to 5.70\%. However, in this case the ACC
slightly drops from 79.31\% to 79.04\%.
Unlike NAD, the full-fledged
ARGD takes the synergy of three losses into account. 
Compared with NAD,  it can 
reduce the ASR from 5.70\% to 2.41\%, while the ACC can be
improved from 79.04\% to 80.38\%.




\section{Conclusion}
\label{sec:con}

This paper proposed a novel 
backdoor defense method named  Attention Relation Graph Distillation (ARGD). Unlike the state-of-the-art  method NAD that only considers attention features of the same order in finetuing and 
distillation,  
ARGD  takes the correlations of attention features with different orders into account. By using our proposed Attention Relation Graphs (ARGs) and corresponding
loss functions,  
ARGD enables quick alignment of 
ARGs between both teacher and student models, thus the impacts of backdoor triggers  can be effectively 
suppressed. Comprehensive experimental results show the effectiveness of our proposed method. 


\bibliographystyle{plain}
\bibliography{ijcai22}

\clearpage
\appendix

\section{Experimental Details}
In this paper,  we implemented six latest backdoor attacks and two latest backdoor defense algorithms.
The backdoor attacks we used are shown as
\begin{itemize}
\item
BadNets. The BadNets trigger is a 3 × 3 checkerboard (pixel values are either 128 or 255) at the top right
corner of images. We labeled the backdoor examples with the target label 0 and achieved a backdoor model 
with a 100\% attack success rate under the backdoor data injection ratio of 20\%.
\item
Trojan attack. We followed the method proposed in the paper \cite{trojan} to reverse engineer a 3 × 3 square
trigger from the last fully-connected layer of the network. We
achieved a backdoor model 
with a 99.81\% attack success rate under the backdoor data injection ratio of 20\%.
\item
Blend attack. We used the random patterns reported in the original paper \cite{blend}. We achieved  a backdoor model with a 79.42\% attack success rate under the backdoor data injection ratio of 20\% and a blend ratio of $\alpha$ = 0.2.
\item
CleanLabel attack. We followed the same settings as reported in the paper \cite{cleanlabel}. We used Projected Gradient Descent (PGD) to generate adversarial perturbations bounded to $L_{\infty }$ maximum perturbation $\varepsilon$ = 0.15.
The trigger is a 3 × 3 grid at the bottom right corner of images.
We achieved a backdoor model 
with a 45.94\% attack success rate under the backdoor data injection ratio of 20\%.
\item
Sinusoidal signal attack (SIG). We generated the backdoor trigger following the horizontal sinusoidal function defined in the original paper \cite{sig} with $\bigtriangleup $ = 20 and $f$ = 6. We achieved a backdoor model with a 99.98\% attack success rate under the backdoor data injection ratio of 20\%.
\item
Reflection attack (Refool). We generated the backdoor reflect image following the generated function defined in the original paper \cite{liu2020reflection}. We achieved a backdoor model with a 100\% attack success rate under the backdoor data injection ratio of 8\%. The open-source code of Refool is available from \\ \textit{https://github.com/DreamtaleCore/Refool}
\end{itemize}

The backdoor defense algorithms we used are shown as follows
\begin{itemize}
\item
Neural Attention Distillation (NAD).
 We used the same experimental setting as reported in their paper for NAD. The open-source code of NAD is available from \textit{https://github.com/bboylyg/NAD}.
 \item
Mode Connectivity Repair (MCR).
We used the open-source code for mode connectivity repair (MCR) and set the endpoint model $t$ = 0 and $t$ = 1 with the same backdoored WRN-16-1. We trained the connection path for 100 epochs and evaluated the defense performance
of the model on the path. Other settings of the code remain unchanged. The open-source code of MCR is available from \textit{https://github.com/IBM/model-sanitization}.
\end{itemize}

\begin{figure}[h]
\centering
\begin{subfigure}{.35\textwidth}
\includegraphics[width=\linewidth]
{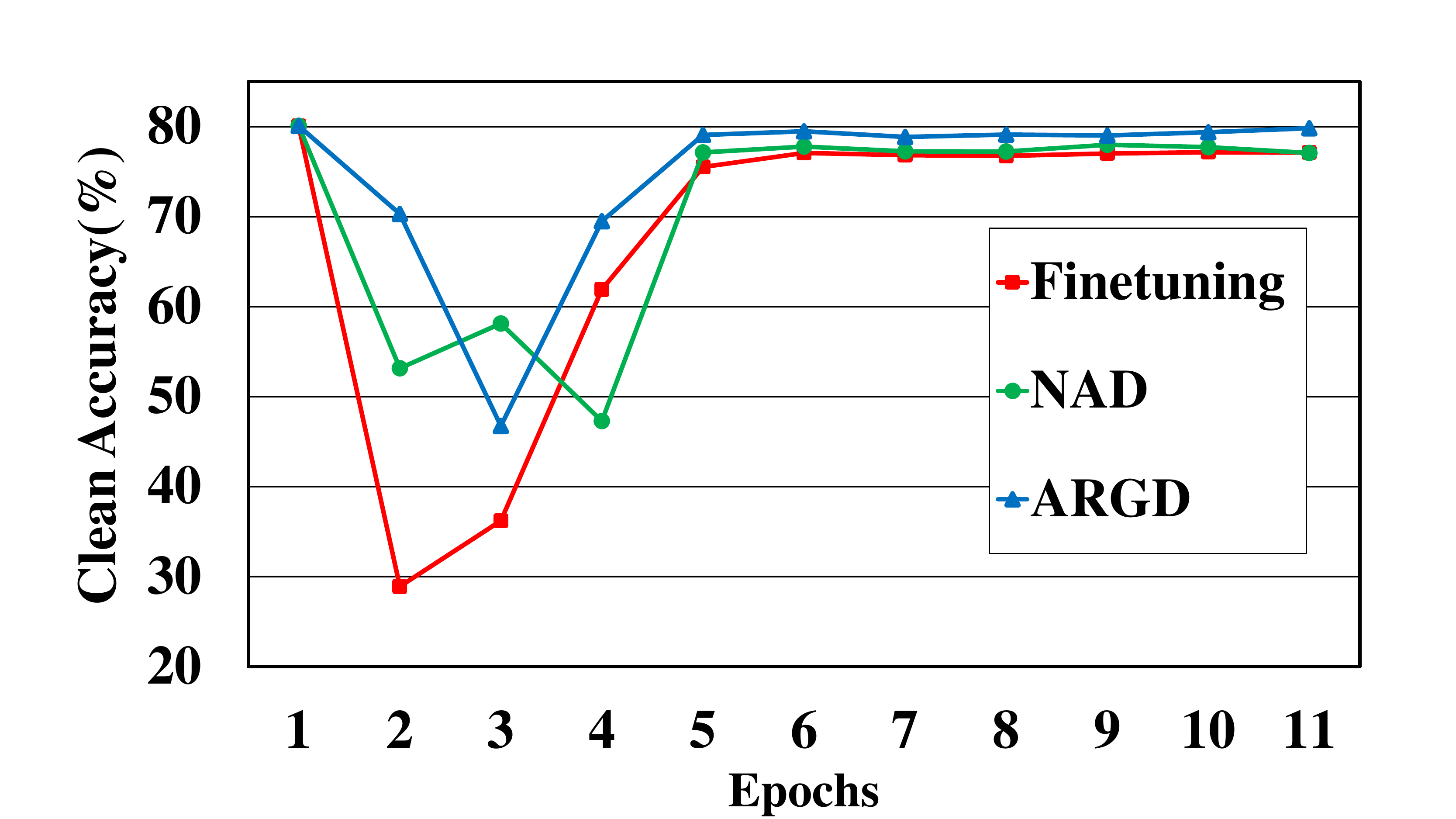}
\caption{BadNets (ACC)}
\end{subfigure}
\hspace{0.15in}
\begin{subfigure}{.35\textwidth}
\includegraphics[width=\linewidth]{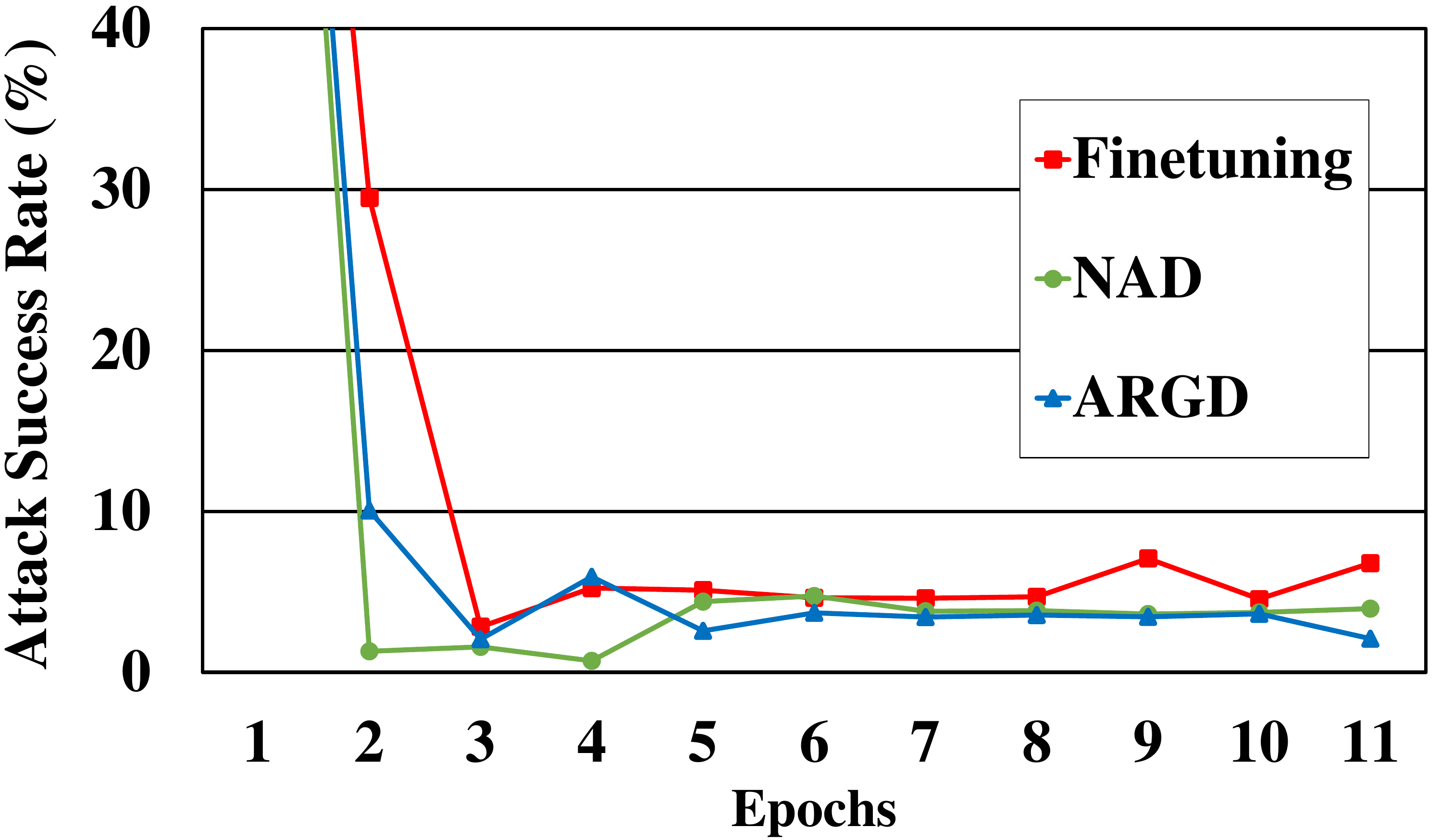}
\caption{BadNets (ASR)}
\end{subfigure}

\begin{subfigure}{.35\textwidth}
\includegraphics[width=\linewidth]{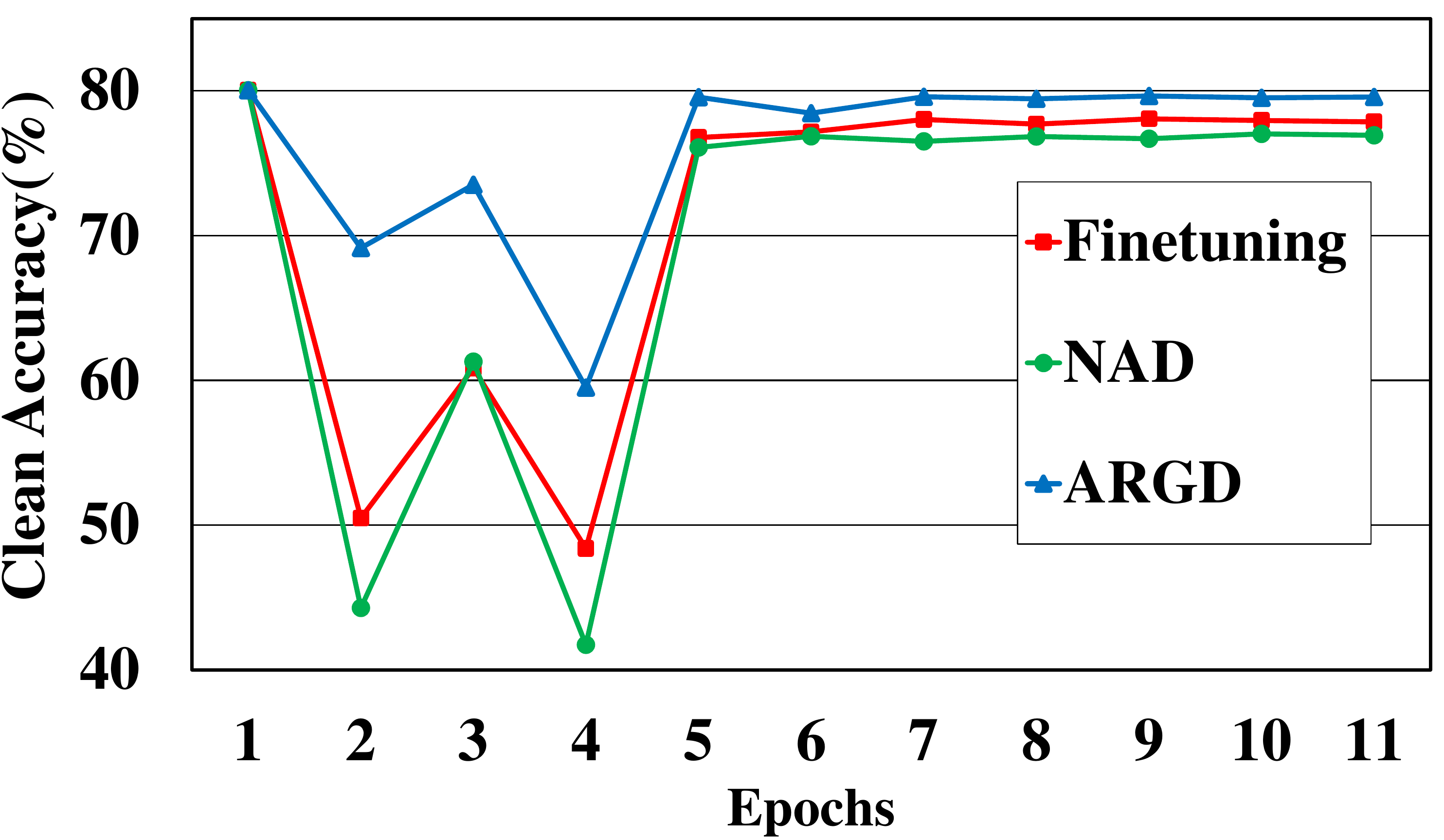}
\caption{Trojan (ACC)}
\end{subfigure}
\hspace{0.15in}
\begin{subfigure}{.35\textwidth}
\includegraphics[width=\linewidth]{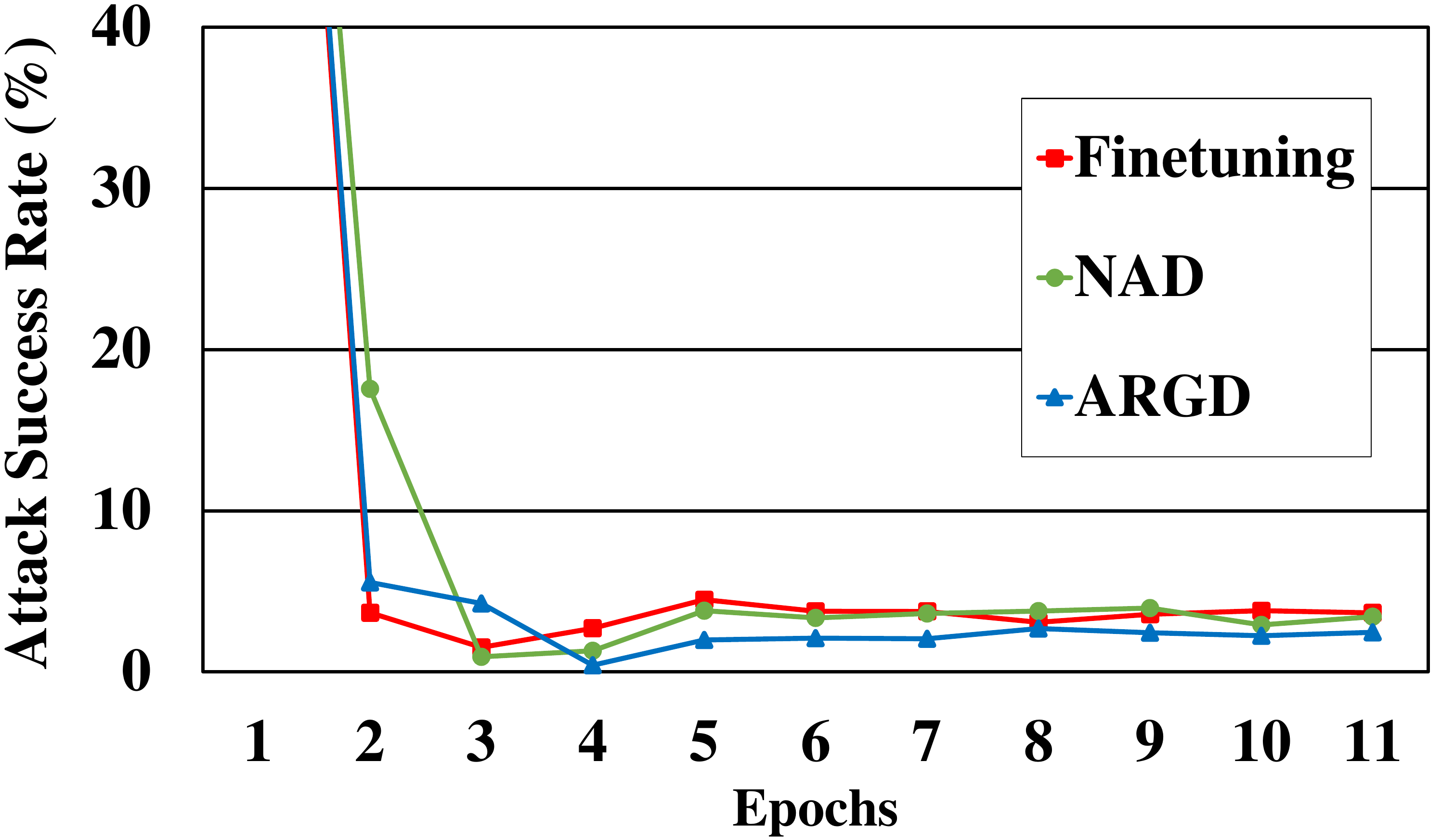}
\caption{Trojan (ASR)}
\end{subfigure}

\begin{subfigure}{.35\textwidth}
\includegraphics[width=\linewidth]{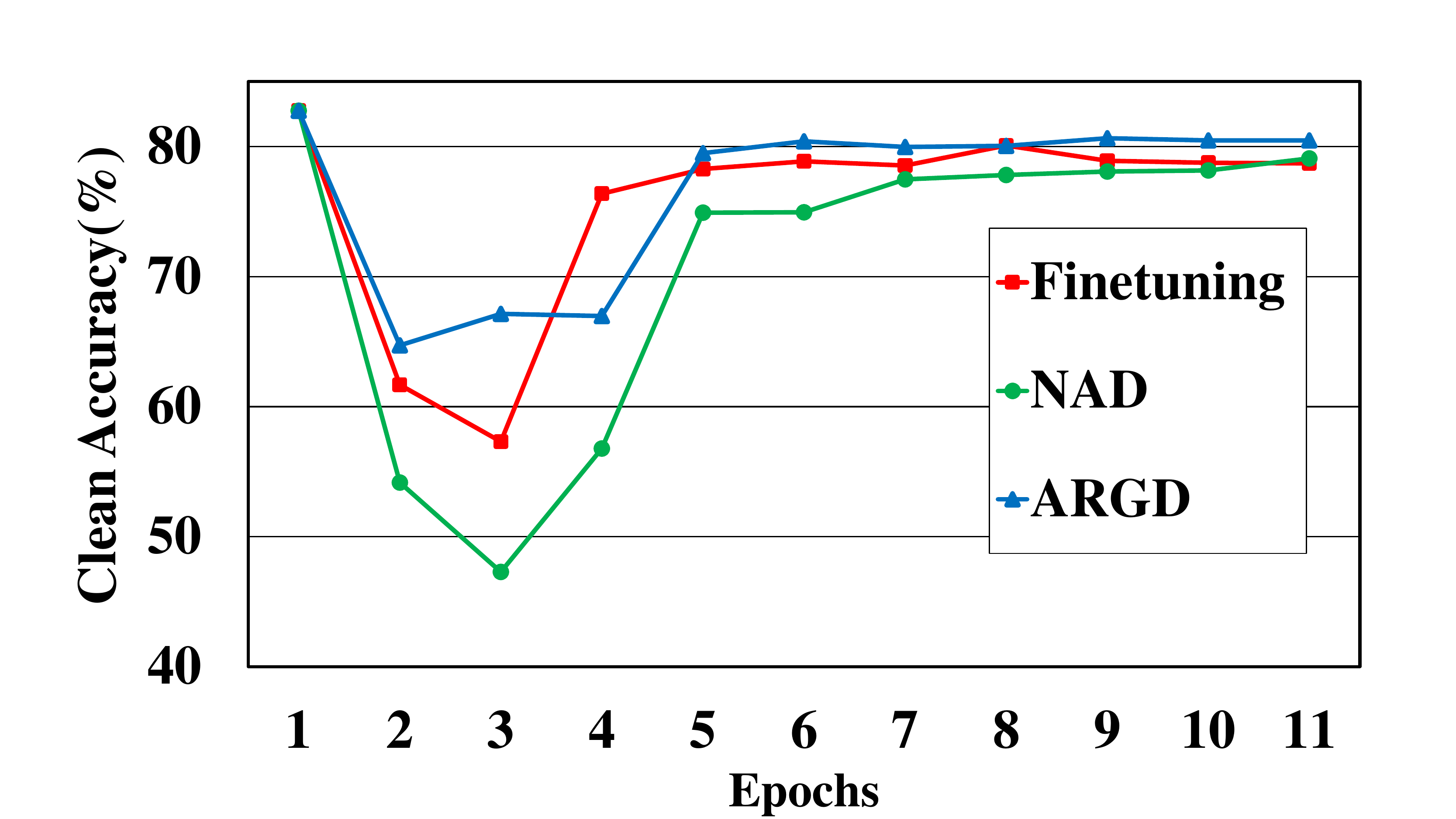}
\caption{Blend (ACC)}
\end{subfigure}
\hspace{0.15in}
\begin{subfigure}{.35\textwidth}
\includegraphics[width=\linewidth]{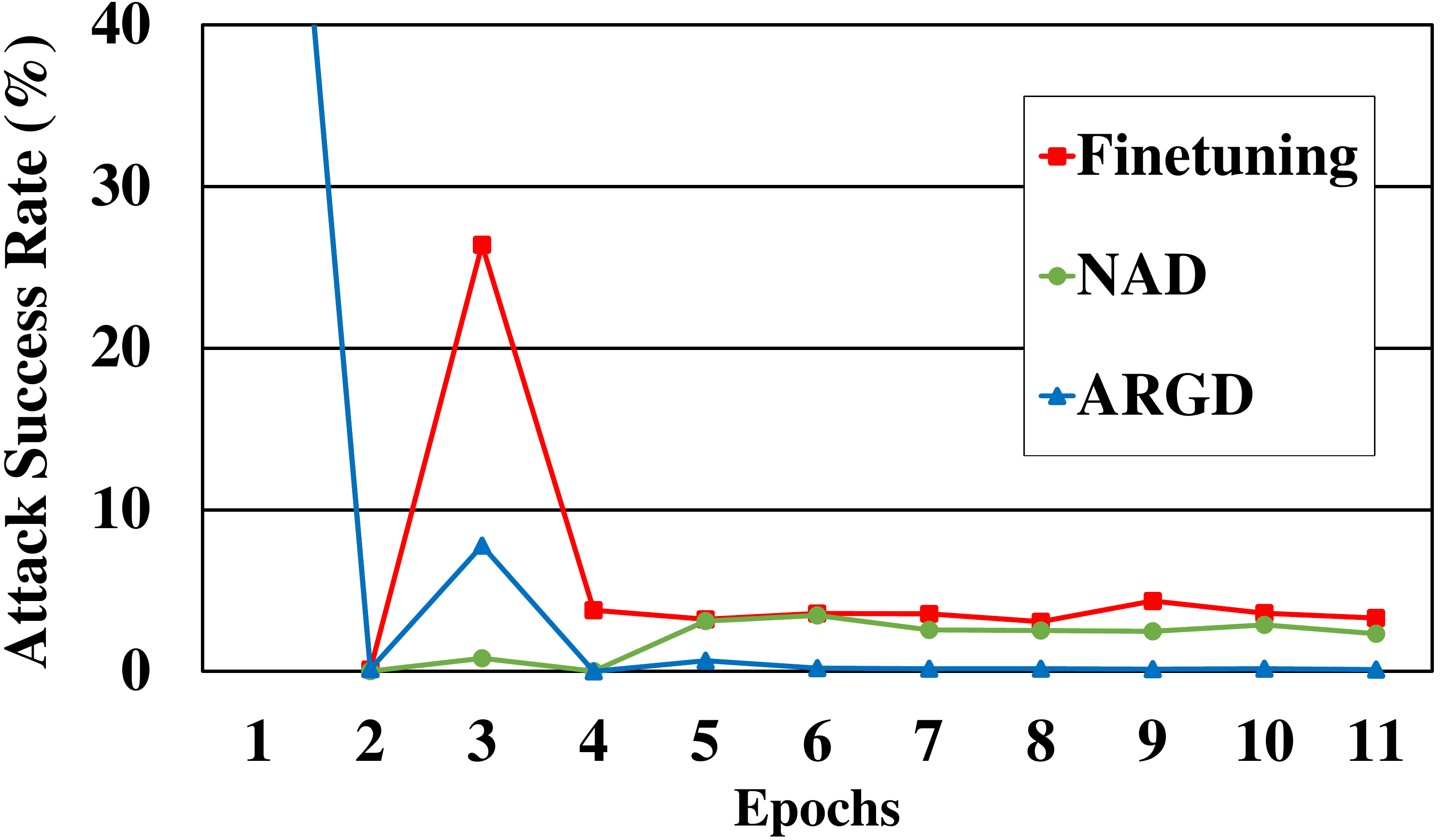}
\caption{Blend (ASR)}
\end{subfigure}

\begin{subfigure}{.35\textwidth}
\includegraphics[width=\linewidth]{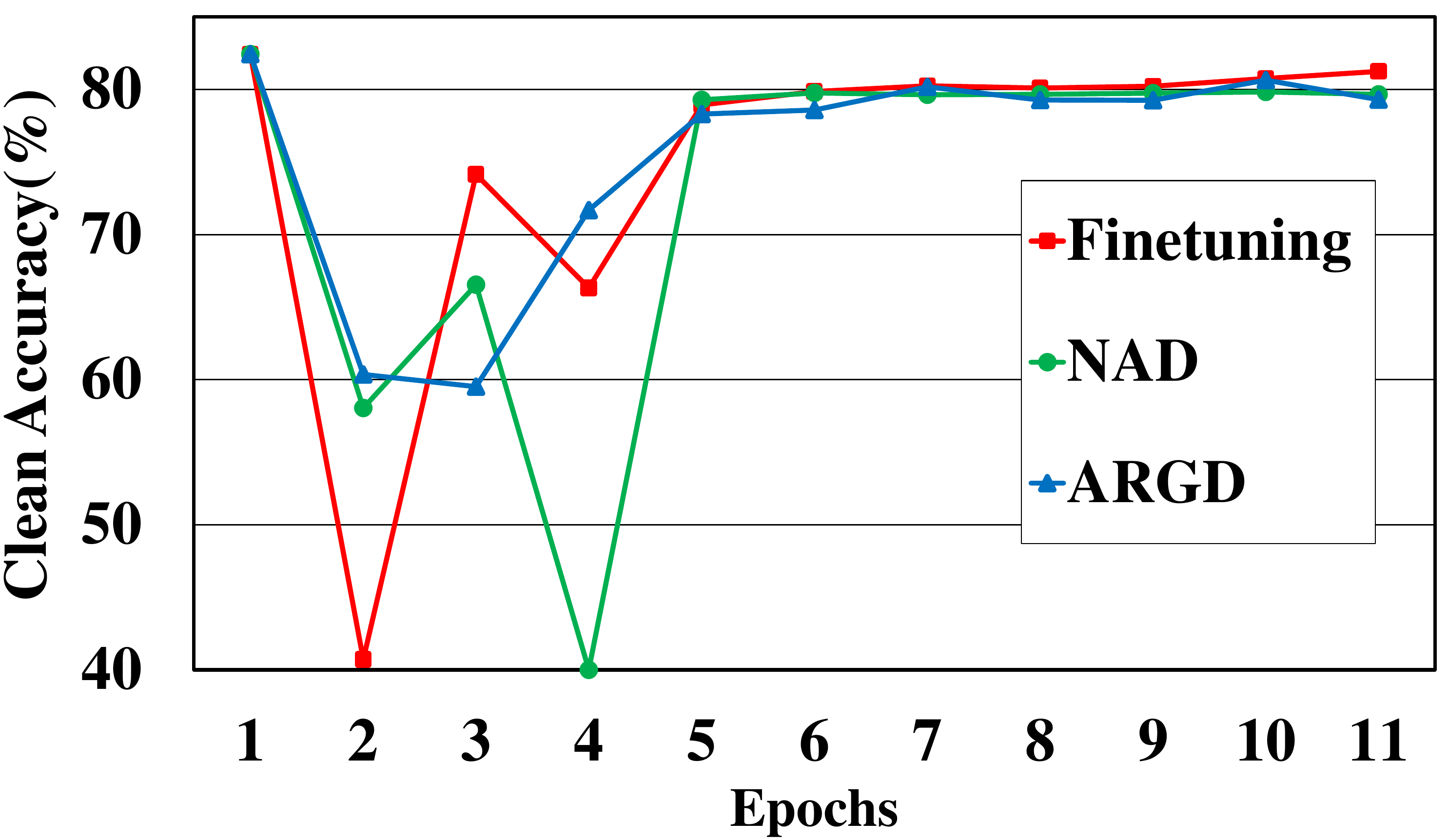}
\caption{CleanLabel (ACC)}
\end{subfigure}
\hspace{0.15in}
\begin{subfigure}{.35\textwidth}
\includegraphics[width=\linewidth]{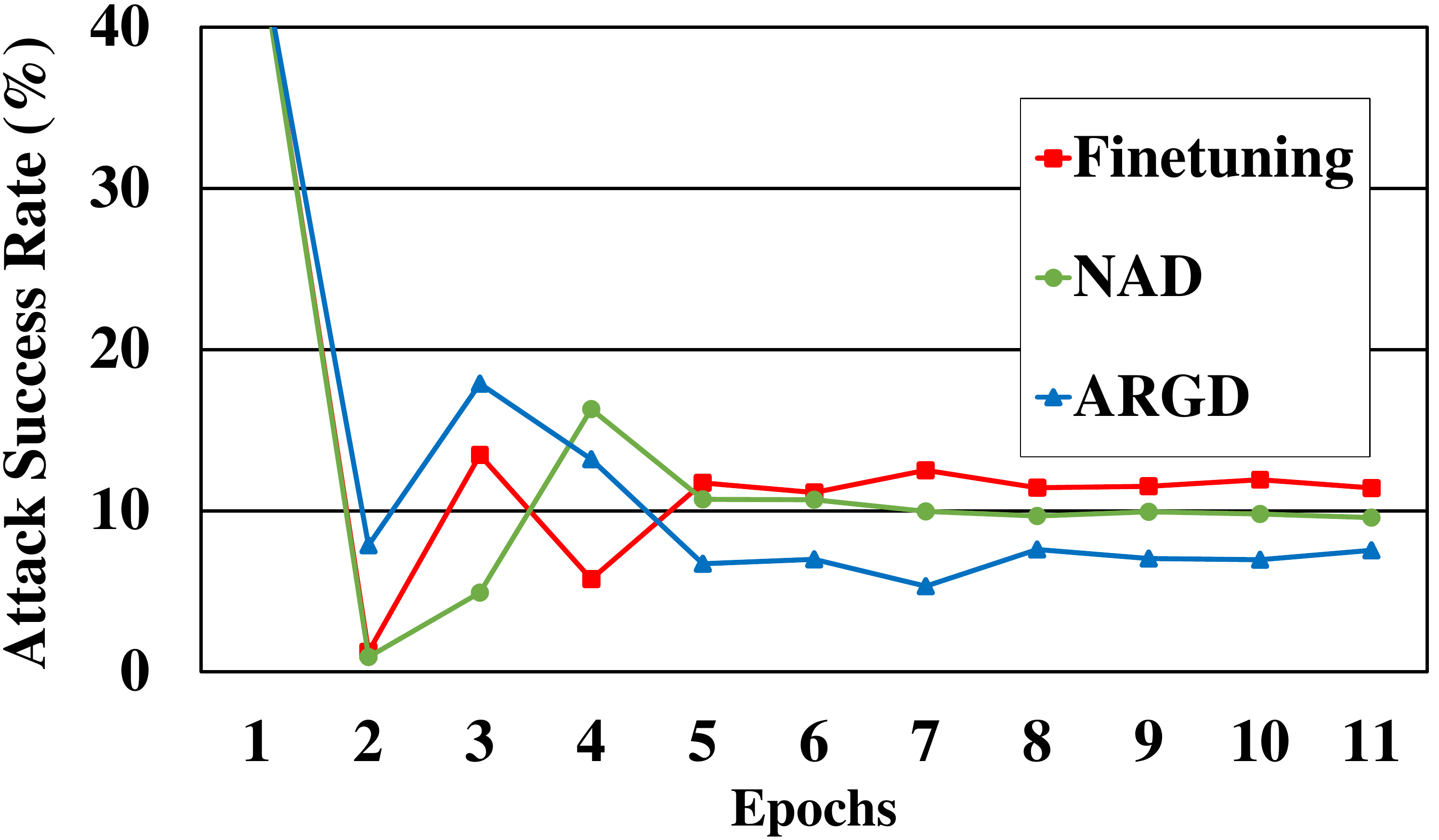}
\caption{CleanLabel (ASR)}
\end{subfigure}

\begin{subfigure}{.35\textwidth}
\includegraphics[width=\linewidth]{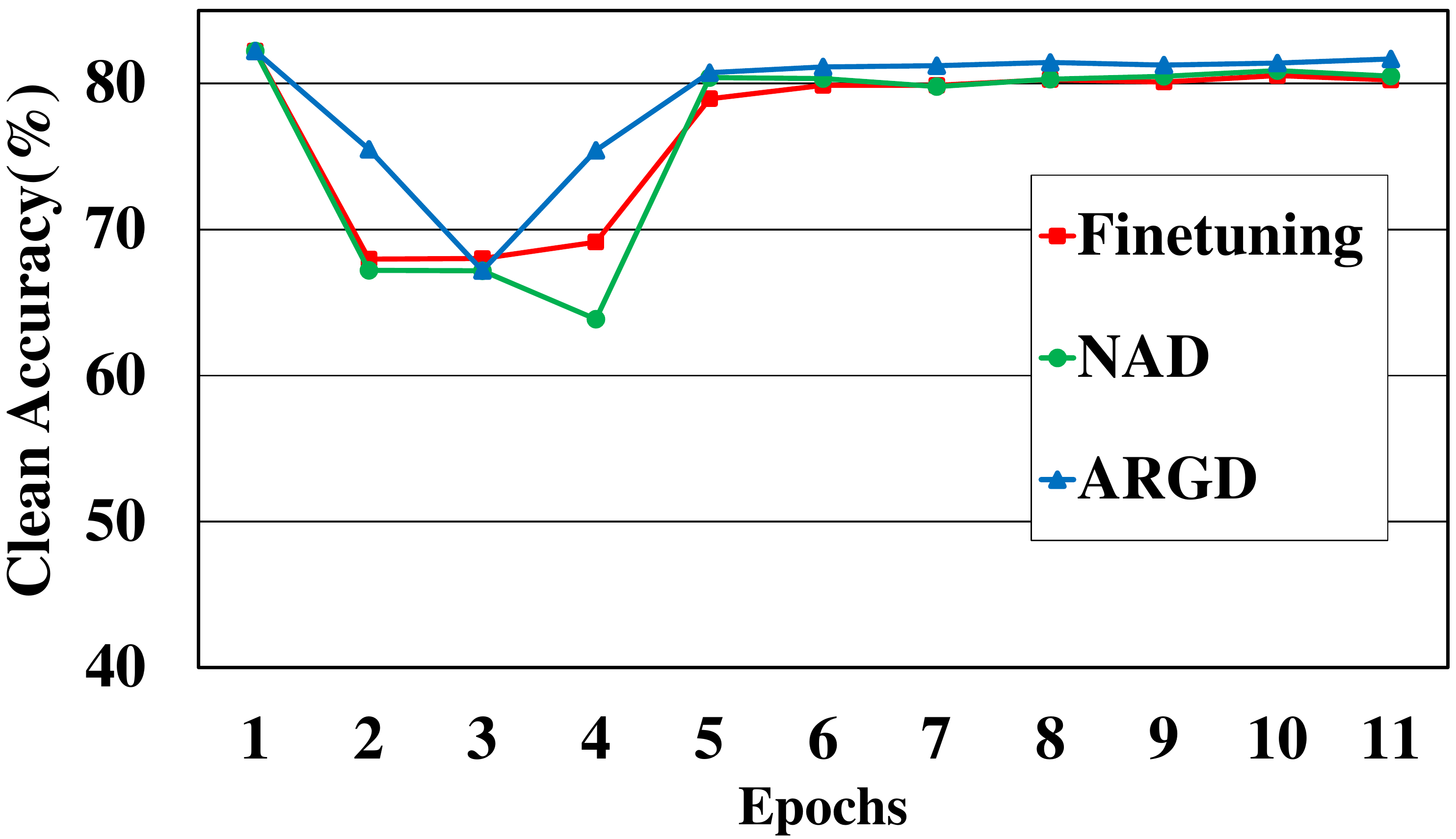}
\caption{Refool (ACC)}
\end{subfigure}
\hspace{0.15in}
\begin{subfigure}{.35\textwidth}
\includegraphics[width=\linewidth]{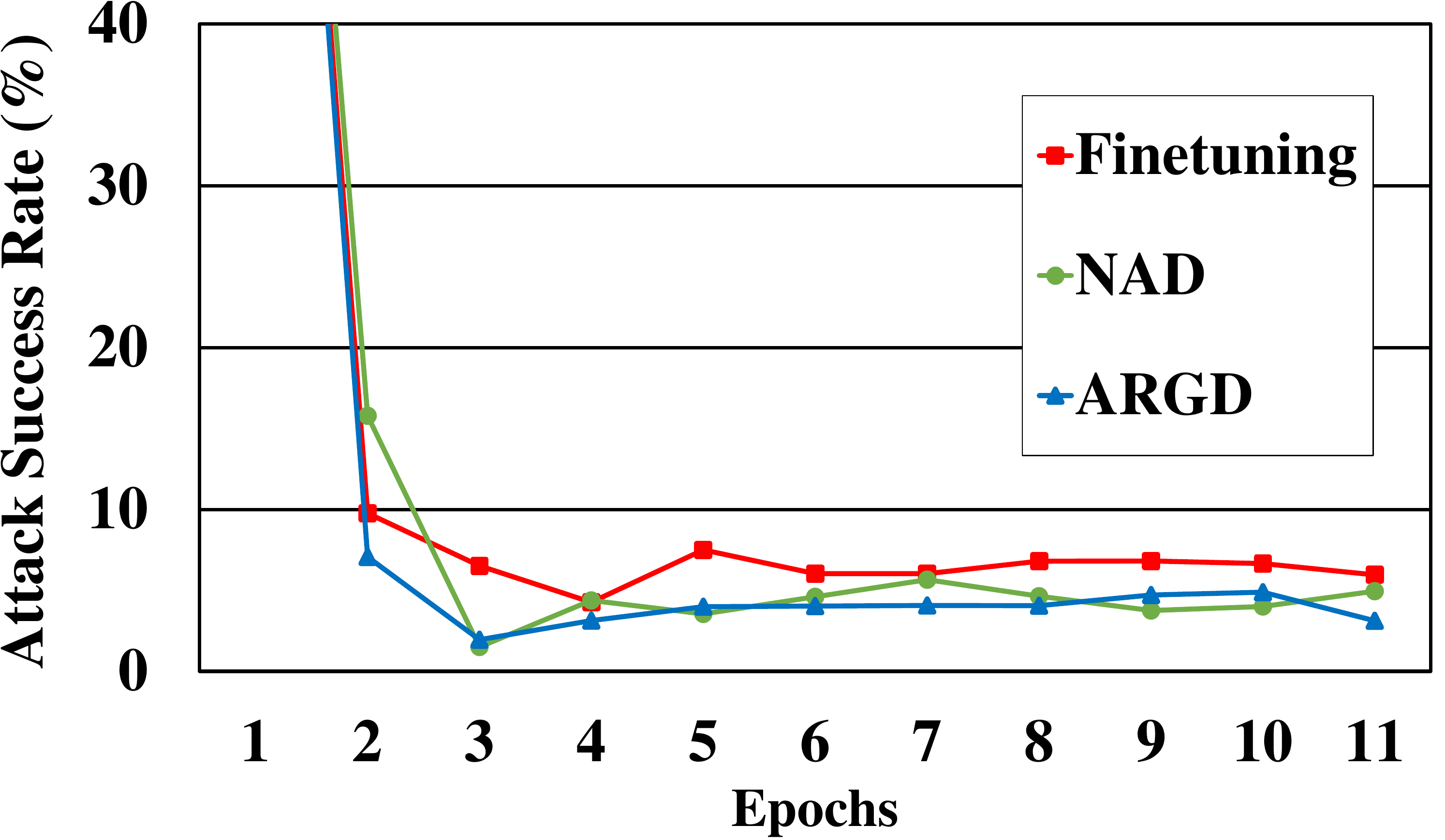}
\caption{Refool (ASR)}
\end{subfigure}

\begin{subfigure}{.35\textwidth}
\includegraphics[width=\linewidth]{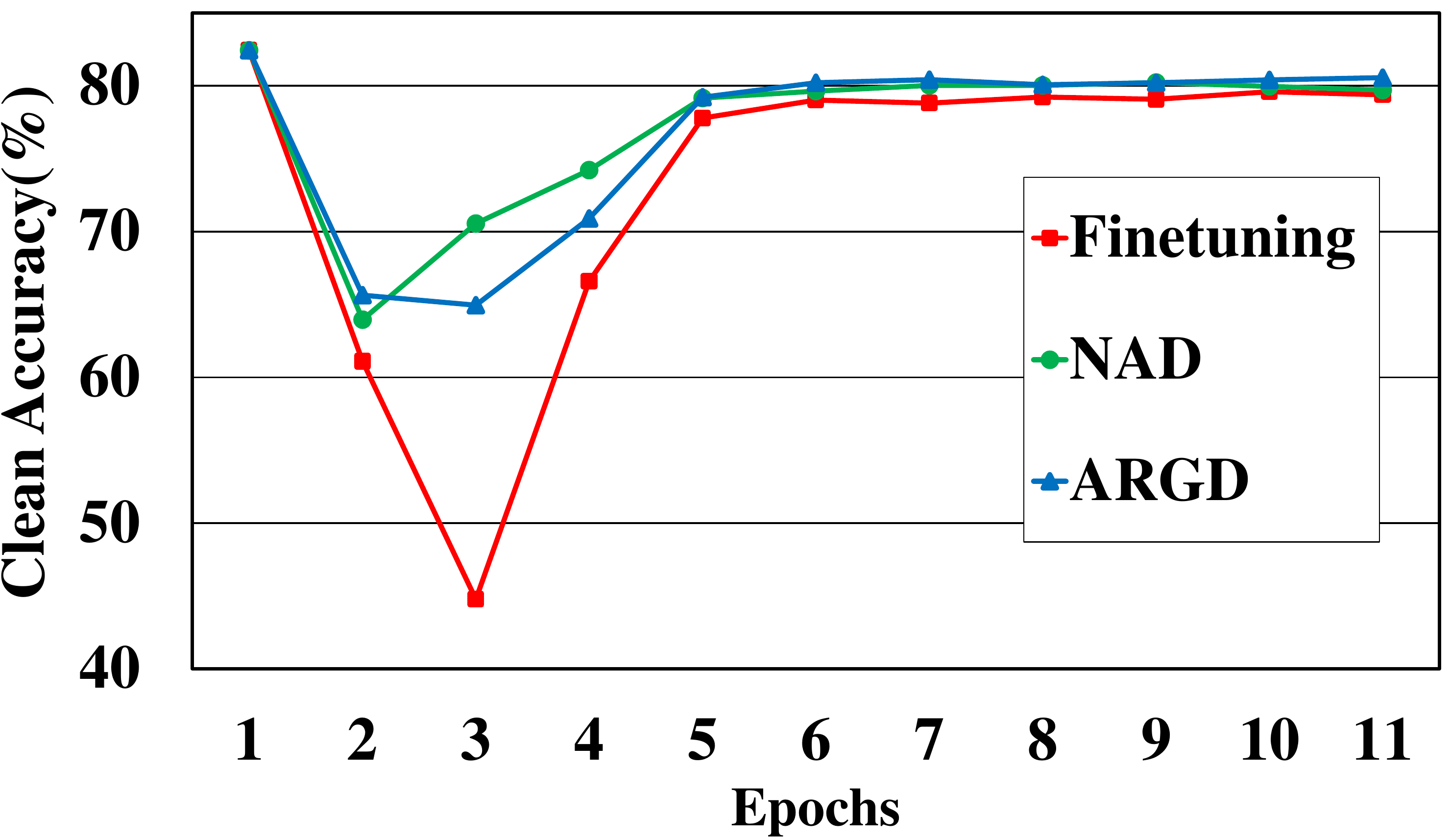}
\caption{SIG (ACC)}
\end{subfigure}
\hspace{0.15in}
\begin{subfigure}{.35\textwidth}
\includegraphics[width=\linewidth]{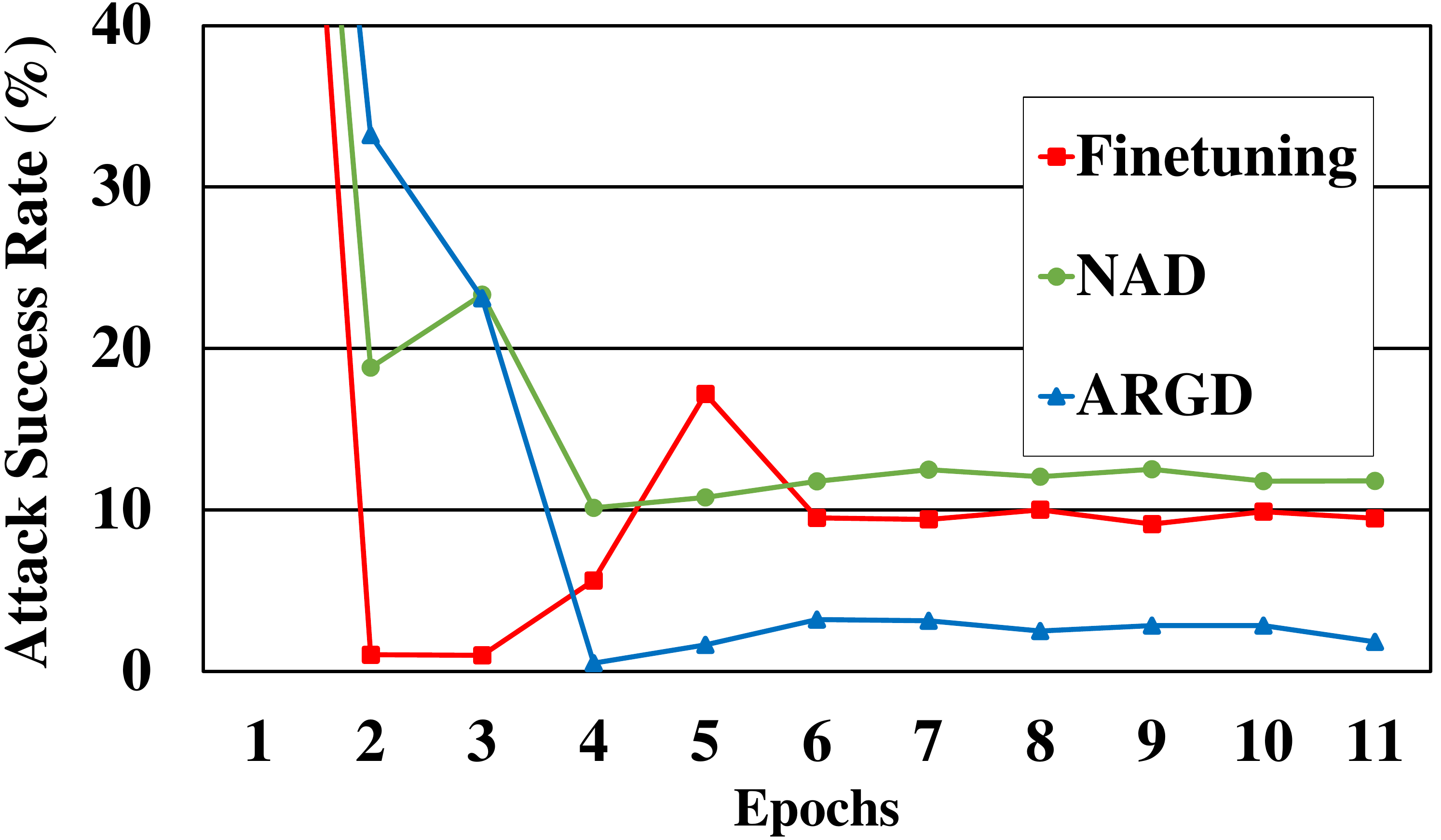}
\caption{SIG (ASR)}
\end{subfigure}
\caption{Convergence rate comparison} 
\label{convergency rate}
\end{figure}
\section{Comparison of Convergence Rate}
In this subsection, we compare the distillation efficiency of our ARGD and the open-source NAD.
Figure
\ref{convergency rate} shows the convergence rate of NAD, finetuning and our ARGD in terms of ASR and ACC against six backdoor attacks within 10 epochs. 
From this figure, we can find that compared with other defense methods, ARGD achieves a faster convergence rate and has the least impact on the ACC of the backdoored DNN. 
For instance, the ASR of the backdoored DNN dramatically reduces to a low level at the $5^{th}$ epoch, while its ACC converges and keeps at a high level. Comparatively, the ACC and ASR of NAD and finetuning cannot converge until the $6^{th}$ or $7^{th}$ epoch.
This convinces the good performance of our ARGD in terms of distillation efficiency achieving a higher convergence rate than NAD and the traditional finetuning method.
Our code is released on \textit{https://github.com/BililiCode/ARGD}.

\end{document}